\UseRawInputEncoding

\documentclass[letterpaper, 10 pt, conference]{ieeeconf} 
\IEEEoverridecommandlockouts
\usepackage{cite}

\usepackage{amsmath,amssymb,amsfonts}
\usepackage{algorithmic}
\usepackage{graphicx}
\usepackage[dvipsnames]{xcolor}
\usepackage{textcomp}
\usepackage{xcolor}
\usepackage{lipsum}  
\usepackage{multirow}
\usepackage{booktabs}
\usepackage{colortbl}
\usepackage{stfloats}
\usepackage{makecell}   % allows multi-line cells easily
\usepackage{subcaption} % put this in your preamble
\usepackage{adjustbox} % for valign
\usepackage{pifont}
\usepackage{circledsteps}
\usepackage{array}      % for m{..}/p{..} columns and newcolumntype
\usepackage[normalem]{ulem}
\newcolumntype{C}[1]{>{\centering\arraybackslash}m{#1}}

\newcolumntype{s}{>{\columncolor[gray]{.90}[.5\tabcolsep]}c}

\def\BibTeX{{\rm B\kern-.05em{\sc i\kern-.025em b}\kern-.08em
    T\kern-.1667em\lower.7ex\hbox{E}\kern-.125emX}}

\definecolor{myBlue}{rgb}{0.1, 0.2, 0.9}

\definecolor{eqnBlue}{HTML}{8FAADC} % for the world model equation
\definecolor{eqnRed}{HTML}{C45812} % for the world model equation
\definecolor{myRed}{RGB}{186, 28, 48}
\definecolor{myDarkGreen}{HTML}{006400} % this is the CSS "DarkGreen"
\definecolor{citecolor}{RGB}{0,100,180} 
\definecolor{darkgreen}{rgb}{0.0,0.5,0.0}

\usepackage[pagebackref=true,breaklinks=true,urlcolor=citecolor,colorlinks,citecolor=citecolor,bookmarks=false,linkcolor=citecolor]{hyperref}

\definecolor{HopBine}{RGB}{170,157,46}
\definecolor{RougeWave}{RGB}{13,82,87}
\definecolor{SolarFlare}{RGB}{211,131,43}
\definecolor{mygray}{RGB}{85,85,85}

\newcommand{\ourcolor}[1]{\textcolor{darkgreen}{#1}}

\newcommand{\ourcolorp}[1]{\textcolor{myBlue}{#1}}

\newcommand{\ours}{\ourcolor{\texttt{decPLM}$\left(\text{const}^+, \text{cf}^{\text{init}}\right)$}}

\newcommand{\oursp}{\ourcolorp{\texttt{decPLM}$\left(\text{const}^+, \text{cf}^+\right)$}}

\newcommand{\base}{\textcolor{orange}{\texttt{decPLM}$\left(\text{const}^-, \text{cf}^{\text{init}}\right)$}}
\newcommand{\basep}{\textcolor{red}{\texttt{decPLM}$\left(\text{const}^-, \text{cf}^+\right)$}}

\begin{document}
% \IEEEsettopmargin{t}{72pt}
% \title{Multi-Quadruped Cooperative Object Transport: \\ Learning Decentralized Pinch-Lift-Move} 
%\title{Decentralized Pinch-Lift-Move: \\
%Learning Scalable Multi-Quadruped Object Transport}
% \author{
%  \IEEEauthorblockN{Anonymous Authors $^1$} % \and
%  }

%%%%%%%%%%%%%%% FOR CAMERA READY %%%%%%%%%%%%%%%%%
\title{\LARGE \bf Multi-Quadruped Cooperative Object Transport: \\ Learning Decentralized Pinch-Lift-Move} 
\author{Bikram Pandit, Aayam Kumar Shrestha, and Alan Fern
% <-this % stops a space
\thanks{*This work is supported by DARPA contract HR0011-24-9-0423 and an NVIDIA Academic Grant. We thank Mohit Gadde and Ashutosh Gupta for their important technical support.}% <-this % stops a space
\thanks{All authors are with Collaborative Robotics and Intelligent Systems Institute (CoRIS), Oregon State University, Corvallis, Oregon, 97331, USA. }
\thanks{Email: \{\footnotesize panditb, aayam.shrestha, afern\}@oregonstate.edu.}
}

\maketitle

% \footnotetext[1]{Oregon State University}

% ╔══════════════════════════════════════════════════════════════════════════════╗
% ║                              SECTION: Abstract                               ║
% ╚══════════════════════════════════════════════════════════════════════════════╝
\begin{abstract}
We study decentralized cooperative transport using teams of $N$-quadruped robots with arm that must pinch, lift, and move ungraspable objects through physical contact alone. Unlike prior work that relies on rigid mechanical coupling between robots and objects, we address the more challenging setting where mechanically independent robots must coordinate through contact forces alone without any communication or centralized control. To this end, we employ a hierarchical policy architecture that separates base locomotion from arm control, and propose a constellation reward formulation that unifies position and orientation tracking to enforce rigid contact behavior. The key insight is encouraging robots to behave as if rigidly connected to the object through careful reward design and training curriculum rather than explicit mechanical constraints. Our approach enables coordination through shared policy parameters and implicit synchronization cues -- scaling to arbitrary team sizes without retraining. We show extensive simulation experiments to 
%validate the approach and
demonstrate robust transport across 2-10 robots on diverse object geometries and masses, along with sim2real transfer results on lightweight objects.

\end{abstract}

\vspace{-0.5em}

% ╔══════════════════════════════════════════════════════════════════════════════╗
% ║                             SECTION: Introduction                            ║
% ╚══════════════════════════════════════════════════════════════════════════════╝

\section{Introduction}

%Object transport within local environments is fundamental to applications in construction, logistics, and emergency response. Legged robots are particularly well-suited for such domains, as they can navigate terrain where wheeled platforms fail~\cite{lee_learning_2020,duan_learning_2023}. However, many real-world objects - boxes, furniture, construction materials, logs, irregular containers - often cannot be easily grasped or rigidly attached to robotic platforms. Recent progress in learning-based loco-manipulation has shown that single humanoid and quadrupedal robots can lift and transport boxes through whole-body coordination~\cite{dao_sim--real_2023, arnold_leva_2025}. These methods succeed when the payload is within the capacity of one robot. However, these objects are often too large or heavy for a single robot, requiring coordinated teams that manipulate them through sustained physical contact. This poses a demanding control problem: How do we design a system where multiple robots can maintain stable contact, distribute forces, and coordinate motion without explicit inter-robot communication and centralized control?

Object transport in local environments is central to construction, logistics, and emergency response. Legged robots are suited for such domains, as they can traverse terrain where wheeled platforms fail~\cite{lee_learning_2020,duan_learning_2023}. Yet many real-world objects---boxes, furniture, construction materials, logs, irregular containers---cannot be easily grasped or rigidly attached. Recent work in learning-based loco-manipulation shows that a single humanoid or quadruped can lift and transport boxes through whole-body coordination~\cite{zhang2025falconlearningforceadaptivehumanoid, dao_sim--real_2023, arnold_leva_2025}, but only for payloads within a single robot's capacity. Larger/heavier objects require coordinated teams manipulating through sustained contact, raising a key question: how can multiple robots maintain stable contact, distribute forces, and coordinate motion without communication or centralized control?

Existing decentralized approaches succeed mainly when robots are rigidly coupled to objects, where mechanical connections passively distribute forces and enforce coordination~\cite{zhang_task-space_2021,vincenti_centralized_2023,turrisi_pacc_2024}. In contrast, manipulating ungraspable objects through contact alone is fundamentally harder: robots must establish and maintain contact, synchronize lifting, and stabilize payloads during motion---without mechanical guarantees or explicit communication. Our key insight is that coordination can still emerge if reward shaping encourages robots to behave \emph{as if rigidly attached} to the payload, even though they are only connected through pinching forces.  

We propose \textbf{decPLM} (\textbf{Dec}entralized \textbf{P}inch-\textbf{L}ift-\textbf{M}ove), a hierarchical control framework with a constellation-based reward that enables quadruped--arm teams to learn robust pinch--lift--move behaviors without communication/centralization. By aligning base and payload motion while maintaining relative poses, $N$-robot teams achieve the benefits of rigid coupling while retaining the flexibility of contact-based manipulation. \emph{Our main contributions are:} \textbf{(1)} a constellation reward unifying position and orientation alignment to enforce rigid-like contact; \textbf{(2)} decentralized training that induces implicit synchronization without communication; \textbf{(3)} evidence that policies trained with small teams ($N{=}2$) generalize to larger teams ($N{=}10$) and diverse payloads. We further analyze when continuous payload tracking is necessary, and demonstrate sim-to-real transfer on physical quadruped--arm robots.

\begin{figure*}[tbh!]
    \centering
    \includegraphics[width=\textwidth]{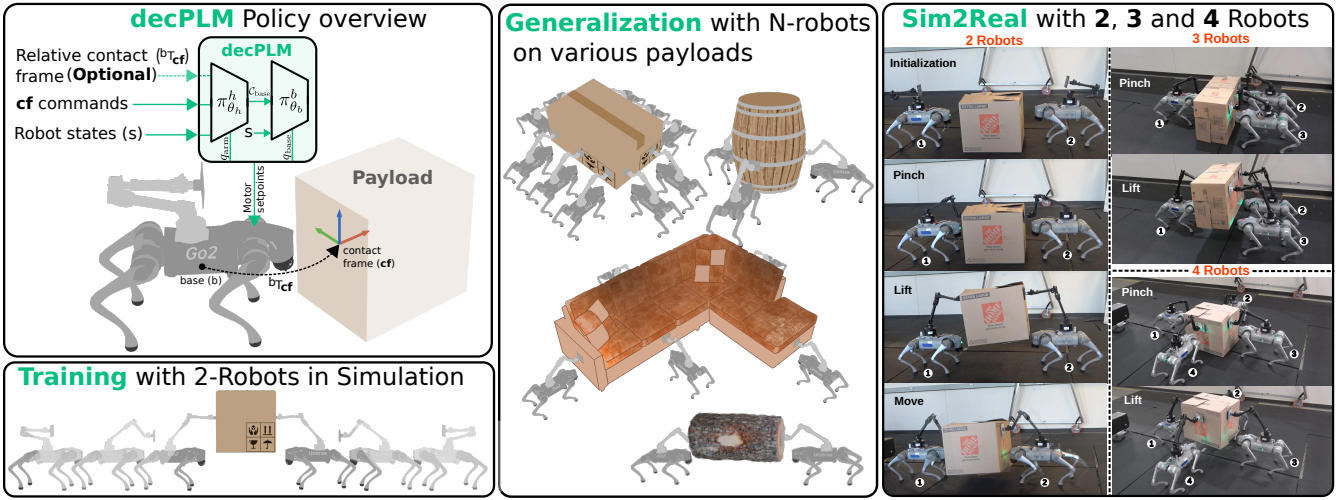}
    \caption{\small Overview of \textbf{decPLM} (\textbf{Dec}entralized \textbf{P}inch-\textbf{L}ift-\textbf{M}ove). 
    \textbf{Top-Left:} Policy structure. Each robot runs the same decentralized policy that receives local proprioception (s), contact-frame command $c_{\text{cf}}$, and (optionally) contact frame pose in base frame ${}^bT_{\text{cf}}$, and outputs arm joint targets and base velocity commands. 
    \textbf{Bottom-Left:} Training setup with two robots in simulation on a box. 
    \textbf{Middle:} Generalization to larger teams and diverse payloads, including a box, a log, a barrel, and a couch.
    \textbf{Right:} Sim2Real Demonstration with 2, 3 and 4 Robots, each running \textbf{\ours{}} policy independently, without any inter-robot communication.}
    \vspace{-1.5em}
    \label{fig:lead_image}
\end{figure*}

% ╔══════════════════════════════════════════════════════════════════════════════╗
% ║                             SECTION: Related Work                            ║
% ╚══════════════════════════════════════════════════════════════════════════════╝

\vspace{-0.5em}
\section{Related Work}
\label{sec:related-work}

We review 3 types of prior work: (1) rigid mechanical coupling between robots and objects, (2) grasping-based manipulation with prehensile contacts, and (3) contact-only manipulation that typically involves planar pushing. In contrast, our approach combines decentralized coordination with contact-only lifting and transport of ungraspable objects.

\textbf{Rigidly Coupled Loco-Manipulation.}
Many successful systems assume mechanical constraints between robots and payloads-through closed chains, fixtures, or passive arms-which greatly simplifies force distribution and stability control~\cite{kennel-maushart_payload-aware_2024,sombolestan_hierarchical_2023,sombolestan_hierarchical_2024,tallamraju_motion_2019,sugar_control_2002}. Centralized approaches use model predictive control and hierarchical planning for mechanically coupled systems~\cite{zhang_task-space_2021,vincenti_centralized_2023}, while passive-arm methods provide stabilization through mechanical compliance~\cite{turrisi_pacc_2024}. Recent work has shown that decentralized control can succeed when robots remain rigidly attached to payloads~\cite{pandit_learning_2025}. In contrast, our approach eliminates mechanical coupling entirely: robots must establish and maintain contact purely through learned policies, changing the coordination problem from force distribution to contact formation and maintenance.

\textbf{
Grasping-Based and Tethered Systems.}
A second category relies on prehensile grasping or non-rigid connections like cables. Multi-robot grasp planning enables coordinated assembly and regrasping but requires reliable graspable features~\cite{dogar_multi-robot_2015}. Cable-towed systems coordinate teams through explicit link dynamics, with taut/slack mode switching that simplifies connection maintenance during transport~\cite{yang_collaborative_2022,chen_decentralized_2025}. While some fully decentralized approaches exist~\cite{farivarnejad_fully_2021}, they still assume reliable attachment mechanisms. These approaches sidestep the core challenge we address: maintaining coordinated contact without mechanical fixtures or tethers.

\textbf{Contact-Only, Decentralized Manipulation.}
Non-prehensile manipulation through contact alone has primarily focused on planar pushing of supported objects, avoiding the challenges of lifting and three-dimensional transport ~\cite{fan_modal_2019,feng_learning_2024,dadiotis_dynamic_2025}. While Single-robot whole-body manipulation has achieved impressive force and position control~\cite{jeon_learning_2023,he_learning_2024,liu_visual_2024,portela_learning_2024}, scaling to cooperative lifting remains unexplored. Moreover multi-agent reinforcement learning has developed relevant coordination mechanisms - hierarchical policies, shared representations, and scalable architectures~\cite{nachum_multi-agent_2020,wu_spatial_2021,an_solving_2024,xiong_mqe_2024,an_scalable_2025} - but has not addressed the specific challenge of maintaining coordinated contact during cooperative lifting. Our approach fills this gap by enabling multiple robots to perform coordinated pinch-lift-move behaviors through contact alone, without mechanical coupling or communication.

% ╔══════════════════════════════════════════════════════════════════════════════╗
% ║                           SECTION: Problem Formulation                       ║
% ╚══════════════════════════════════════════════════════════════════════════════╝

\section{Problem Formulation}
\label{sec:problem-formulation}
We consider decentralized cooperative transport where a team of $N$ quadrupedal robots must transport a single rigid-body object, which we refer to as the \emph{payload}. The payload is assumed to be ungraspable, that is, it cannot be rigidly attached to any robot and must instead be manipulated through physical contact at its surfaces using contact pads attached to each robot's arm. The robots must coordinate to \emph{pinch}, \emph{lift}, and \emph{move} the payload so that it tracks a commanded motion trajectory of the payload.

In this paper, we focus on the decentralized pinch-lift-move (PLM) control problem and assume a higher-level planner that provides: (1) contact frames on the payload, one per robot, that allows force closure; (2) a synchronized start signal to initiate the collaborative pinch and lift maneuver; and (3) velocity commands for the payload transformed into contact-frame coordinates for each robot. The planner specifies only geometrically feasible contact locations and transport commands. Contact maintenance, force regulation, synchronization, and dynamic stability during transport are not enforced by the planner and must be achieved through decentralized control. We assume that each pinch-lift-move episode begins with robots positioned within a specified tolerance of their contact frames.

\subsection{Technical Formulation}

We assume a team of $N$ quadruped robots, each equipped with a single arm whose end-effector is a flat contact pad. The $N$-robot PLM problem is specified through $N$ contact frames $\{{}^{b}T^{(r)}_{\text{cf}}\}_{r=1}^{N}$, where each contact frame 
${}^{b}T^{(r)}_{\text{cf}} \in SE(3)$\footnote{Represented in the observation 
as a 7D vector (position and quaternion).} specifies a surface location on the payload (relative to the base) 
where robot $r$ should engage its contact pad.

Once the payload has been lifted, its motion is driven by an instantaneous team-level payload transport command
% The overall transport objective is expressed by a high-level object command
\(
\mathcal{C}_{\text{pl}} = \big(v_{\text{pl}},\, \omega_{\text{pl}},\, h_{\text{pl}}\big),
\)which specifies joystick-like inputs for planar velocity, yaw rate, and height of the payload's root frame.
% \[
% \mathcal{C}_{\text{pl}} = \big(v_{\text{pl}},\, \omega_{\text{pl}},\, h_{\text{pl}}\big),
% \] 
% which specifies joystick-like inputs for planar velocity, yaw rate, and height of the payload’s root frame.
Because the team-level command $\mathcal{C}_{\text{pl}}$ depends on the payload's geometry, which is not known to the robots, it is converted into per-robot contact-frame commands
\(
\mathcal{C}^{(r)}_{\text{cf}} = \big(v^{(r)}_{\text{cf}},\, \omega^{(r)}_{\text{cf}},\, h^{(r)}_{\text{cf}}\big).
\)
Specifically, for robot $r$ with contact frame at fixed offset $p^{(r)}_{\text{offset}}$ from the payload root (where $\mathcal{C}_{pl}$ is defined), we compute:
% \begin{align*}
% v^{(r)}_{\text{cf}} &= v_{\text{pl}} + \omega_{\text{pl}} \times p^{(r)}_{\text{offset}} \\
% \omega^{(r)}_{\text{cf}} &= \omega_{\text{pl}} \\
% h^{(r)}_{\text{cf}} &= h_{\text{pl}} + p^{(r)}_{\text{offset},z}
% \end{align*}
\vspace{-1.5em}

\begin{small}
\[
v^{(r)}_{\text{cf}} = v_{\text{pl}} + \omega_{\text{pl}} \times p^{(r)}_{\text{offset}};\;\;\; \omega^{(r)}_{\text{cf}} = \omega_{\text{pl}};\;\;\;
h^{(r)}_{\text{cf}} = h_{\text{pl}} + p^{(r)}_{\text{offset},z}
\]
\end{small}

%
%Here $p^{(r)}_{\text{offset}}$ is a constant vector determined solely by the rigid-body geometry of the object 
%and the placement of the contact frames. 
Importantly, the offsets $p^{(r)}_{\text{offset}}$ do not change with motion, 
so the transformation from $\mathcal{C}_{\text{pl}}$ to $\mathcal{C}^{(r)}_{\text{cf}}$ is 
state-free, inexpensive to compute, and identical across time.

% ╔══════════════════════════════════════════════════════════════════════════════╗
% ║                           SECTION: System Overview                           ║
% ╚══════════════════════════════════════════════════════════════════════════════╝

\section{System Overview}

Cooperative transport requires a team of $N$ quadruped-arm robots to perform coordinated \emph{pinch}, \emph{lift}, and \emph{move} operations. Fundamentally, this is a high-dimensional control problem where each robot must coordinate all leg and arm joints to maintain locomotion while regulating contact forces on the payload. To simplify this challenge, we adopt a hierarchical policy architecture for each robot that builds on well-established low-level locomotion control for quadrupeds.
%The cooperative transport task requires a team of $N$ quadruped-arm robots to perform coordinated \emph{pinch}, \emph{lift}, and \emph{move} operations. 
%We decompose this into two control channels: (1) base locomotion for positioning and transport, and (2) arm manipulation for contact establishment and force regulation. This natural decomposition motivates our hierarchical policy architecture.

\textbf{Hierarchical Policy Architecture:} We employ a two-level hierarchy that separates locomotion from manipulation.

\emph{(1) Low-Level Policy.} The base locomotion policy $\pi^b_{\theta_b}$ provides a velocity-control abstraction to the high-level policy, where $\theta_b$ are the learned policy parameters. The input to $\pi^b_{\theta_b}$ is the proprioceptive state $s^{(r)}$ and desired base velocity commands $\mathcal{C}_{\text{base}}^{(r)} = (v_{\text{base}}^{(r)}, \omega_{\text{base}}^{(r)})$ of robot $r$, specifying the target linear and angular velocities. The output is the target joint positions $q_{\text{base}}^{(r)}$ for the quadruped base, which are then tracked by PD controllers to produce motor torques.

\emph{(2) High-Level Policy.} As illustrated in Figure \ref{fig:lead_image}, the high-level policy $\pi^h_{\theta_h}$ coordinates task-level behavior. Its inputs when used for robot $r$ include: proprioceptive state $s^{(r)}$, contact-frame pose ${}^{b}T_{\text{cf}}^{(r)}$ (optionally available) in the robot's base frame, contact-frame command $\mathcal{C}_{\text{cf}}^{(r)}$, and a temporal synchronization signal $t^{\text{sync}} \in [0,5]$ that provides weak coordination cues during initial contact formation. In particular, when the policy is started before pinching $t^{\text{sync}}=0$ and then increments for 5 seconds (see Section \ref{sec:training} for details). 

Due to the challenges of accurate tracking of the contact-frame pose ${}^{b}T_{\text{cf}}^{(r)}$, we consider two modes of operation: (1) the contact frame pose is continuously available, and (2) the contact frame pose is only available at the initial timestep and then masked thereafter. In the latter case, the robot must rely on proprioception to infer information about the contact frame for effective force regulation.

\label{subsec:decen-policy-spec}
\textbf{Decentralized Policy Execution:} At execution time, each robot independently runs $\pi^h_{\theta_h}$ and $\pi^b_{\theta_b}$ at 50\,Hz using only local information. 
First, the high-level policy produces arm joint targets and a base command:
\begin{small}
\[
(q_{\text{arm}}^{(r)},\; \mathcal{C}_{\text{base}}^{(r)}) 
= \pi^h_{\theta_h}\left(s^{(r)},\; {}^{b}T_{\text{cf}}^{(r)},\; \mathcal{C}_{\text{cf}}^{(r)},\; t^{\text{sync}}\right).
\]
\end{small}
Next, the low-level policy maps the base command into joint targets:
\begin{small}
\(
q_{\text{base}}^{(r)} 
= \pi^b_{\theta_b}\left(s^{(r)},\; \mathcal{C}_{\text{base}}^{(r)}\right).
\)
\end{small}
Each robot then sends 
\begin{small}
\(
\left(q_{\text{arm}}^{(r)}, q_{\text{base}}^{(r)}\right)
\)
\end{small}
to the joint-level PD controllers. 

Robots are initialized near their respective payload contact frames. 
A synchronized start signal sets $t^{\text{sync}}=0$, initiating the pinch-lift phase. 
After $t^{\text{sync}}=5$, this phase ends and $\pi^h_{\theta_h}$ begins receiving transport (move) commands.

% The key inputs include: proprioceptive state $s^{(r)}$, contact-frame pose ${}^{b}T_{\text{cf}}^{(r)}$ (optionally available) in robot base frame, contact-frame command $\mathcal{C}_{\text{cf}}^{(r)}$, and temporal synchronization signal $t^{\text{sync}} \in [0,5]$ that provides weak coordination cues during initial contact formation.
 
This architecture learns complex coordination behaviors - contact placement, force regulation, and lift synchronization - in the high-level layer while ensuring reliable locomotion through the low-level controller. Coordination emerges through shared parameters, common objectives, and physical interaction, without requiring explicit communication.

% ╔══════════════════════════════════════════════════════════════════════════════╗
% ║                           SECTION: Training Approach                         ║
% ╚══════════════════════════════════════════════════════════════════════════════╝

\section{Training Approach}
\label{sec:training}

The base velocity controller $\pi^b_{\theta_b}$ is pre-trained using standard locomotion rewards and PPO~\cite{schulman_proximal_2017}, following established approaches for legged locomotion~\cite{noauthor_closing_2024}. We use a 3-layer 128x128x128 MLP with ELU units as the policy architecture. Details can be found in the extended paper. The pre-trained $\pi^b_{\theta_b}$ is frozen during high-level policy training. 

For the high-level policy $\pi^h_{\theta_h}$, we use Multi-Agent Proximal Policy Optimization (MAPPO)~\cite{yu_surprising_2021} within the Centralized Training and Decentralized Execution (CTDE)~\cite{amato_introduction_2024} paradigm. We also explored IPPO ~\cite{dewitt2020independentlearningneedstarcraft} training without centralized critics. In this setting, learning was unstable due to non-stationarity introduced by multiple simultaneously adapting agents, and policies failed to achieve synchronized lifting or stable transport. We therefore adopt CTDE to stabilize coordination learning. Training can be conducted with any number of robots, with update time scaling roughly linearly in $N$. Surprisingly, as our results show, training with only $N=2$ robots is sufficient to produce policies that generalize well to much larger teams.

During training, we use asymmetric critic networks \cite{pinto_asymmetric_2018}, which include privileged information that is unavailable to the actor. In particular, each robot maintains its own critic network that observes both local inputs available to the actor and the local inputs of all other robots to provide richer learning signals. At execution time, robots use only the shared policy $\pi^h_{\theta_h}$ with local observations, ensuring fully decentralized operation for an arbitrary number of robots.

The policy $\pi^h_{\theta_h}$ is implemented as a 2-layer 128x128 MLP that concatenates the local observation vector as indicated in~\ref{subsec:decen-policy-spec} with the previous action (policy output). This action history provides important temporal context for maintaining contact stability, particularly in the execution mode where contact frame pose information is masked out. All robots share policy parameters $\theta_h$ while maintaining separate critic networks during training. We assume homogeneous robots and shared policy parameters; heterogeneity in contact geometry arises through randomized contact frame assignments rather than distinct policy roles.

All policies are trained in the IsaacLab~\cite{mittal_orbit_2023} simulator with 2048 parallel environments, each containing a team of $N$ Unitree Go2 quadrupeds equipped with Unitree D1 arms that jointly manipulate a box. Training builds on the RSL-RL framework~\cite{rudin_learning_2022}. The following subsections detail two key design choices that critically affect learning outcomes: the reward function and the training curriculum.
%where we implement Multi-Agent PPO (MAPPO)~\cite{yu_surprising_2021} under the CTDE~\cite{amato_introduction_2024} paradigm.

\vspace{-0.5em}
\subsection{Reward Design}
\label{sec:reward-design}

\begin{figure}[t]
    \centering
    \includegraphics[width=0.8\columnwidth]{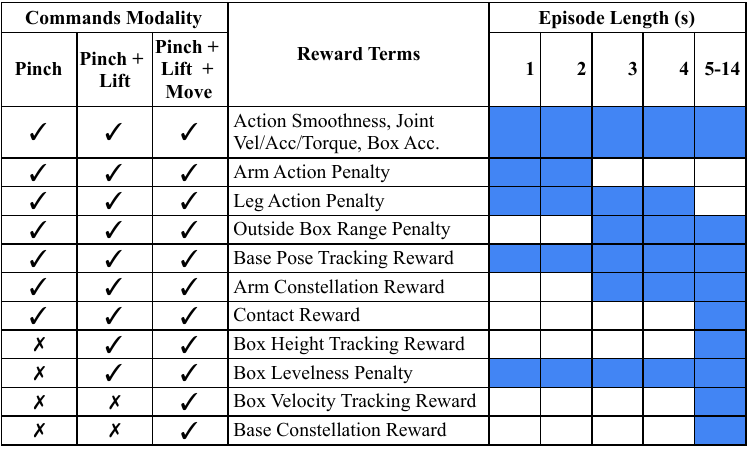}
    \caption{\small Reward activation schedule: Each row lists a reward term along with the command phase (pinch, lift, move) in which it is used. The blue bars show when, within an episode, each reward is active. This illustrates how different rewards are introduced and shaped across both training phases and episode time.}
    \vspace{-1.2em}
    \label{fig:reward_timeline}
\end{figure}

Our reward function addresses the challenge of learning rigid contact behavior and coordinated motion without mechanical constraints. A novelty of our design is a \emph{constellation reward} which provides a unified supervision for contact formation and velocity tracking. This is supplemented by more traditional task-specific tracking rewards and auxiliary penalties that stabilize training. Fig~\ref{fig:reward_timeline} shows the reward components and when they are activated across training phases and episode time (phase/time details in Sec. \ref{sec:training_curriculum}). 

\subsubsection{Constellation Reward}

\begin{figure}[t]
    \centering
    \includegraphics[width=0.9\columnwidth]{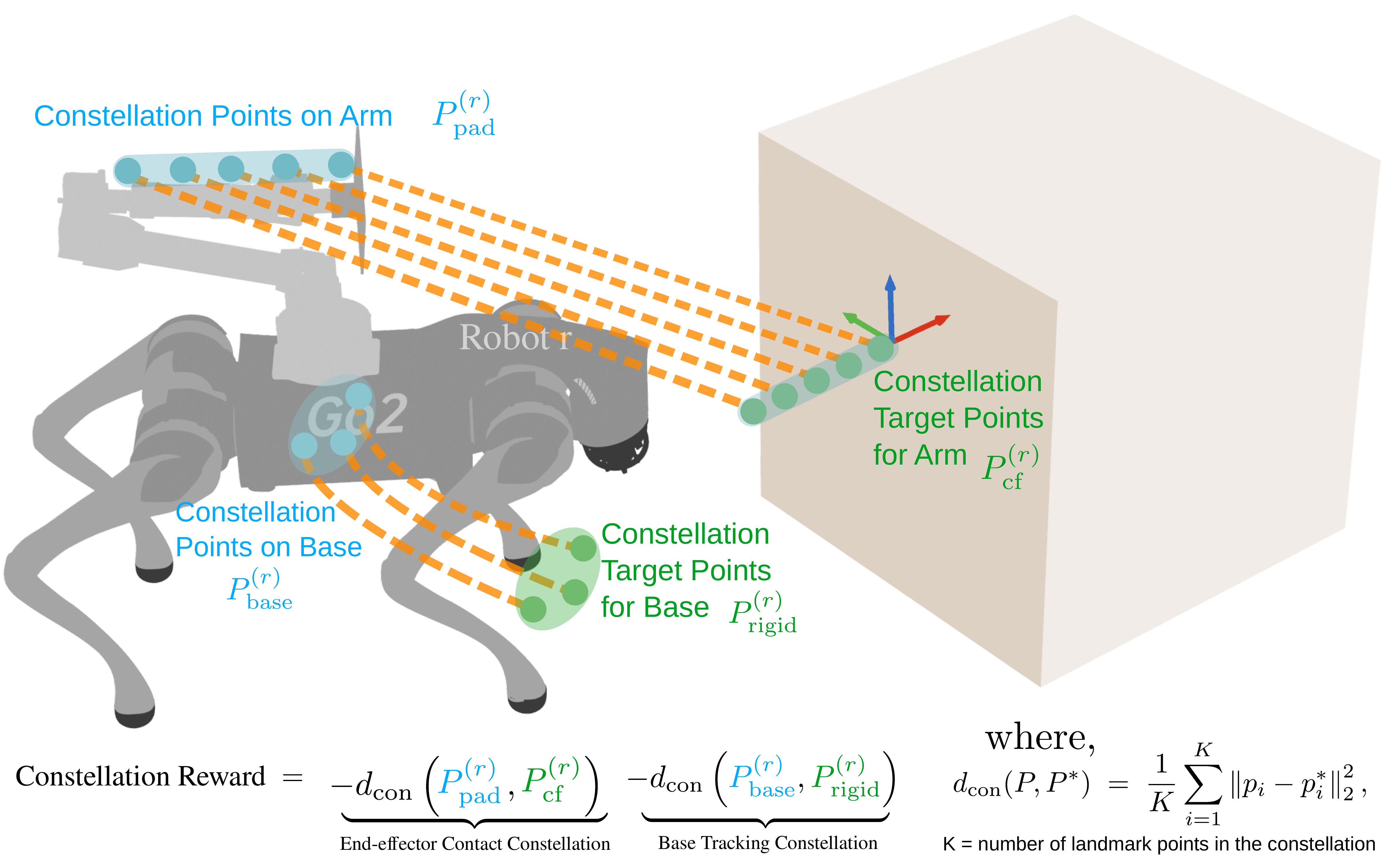}
    \caption{\small Constellation reward illustration. 
    Source points \textcolor{cyan}{(blue)} on the pad and base are matched to their corresponding target points \textcolor{darkgreen}{(green)}. 
    Dotted \textcolor{orange}{orange lines} show the errors that the policy must minimize, with the two groups representing 
    End-effector Contact Constellation and Base Tracking Constellation.}
    \vspace{-1em}
    \label{fig:constellation-reward}
\end{figure}

% The \emph{constellation reward} provides a unified mechanism to enforce both positional and orientational alignment between robot and contact frames. The key idea is that successfully learning to maintain these alignments results in an overall multi-robot system that behaves as if they are rigidly attached. 

The \emph{constellation reward} provides a unified mechanism for enforcing both positional and orientational alignment between each robot and its designated contact frame. We define two constellations of points, one anchored to the robot's base and one to arm end-effector, each paired with a corresponding target constellation anchored to the payload. This constellation-based approach is inspired by point-set registration methods in computer vision~\cite{besl_method_1992, chen_object_1992, arun_least-squares_1987}, where both translation and rotation are implicitly encoded through the alignment of two point sets. The intuition is that by maintaining these, the team behaves as if the robots were rigidly attached to the payload.

Formally, we represent a constellation as a set of landmark points $P=\{p_1, \ldots, p_K\}$ rigidly anchored to a reference frame, and define a corresponding target constellation $P^*=\{p_1^*, \ldots, p_K^*\}$ anchored to the payload. The constellation distance is then defined as the mean squared error between matched landmarks:
\begin{small}
\[
d_{\text{con}} (P,P^*)\;=\; \frac{1}{K} \sum_{i=1}^{K} \left\| p_i - p_i^{*} \right\|_2^2,
\label{eq:constellation_loss}
\]
\end{small}
which provides a smooth signal that couples position and orientation through the geometry of the point set. 

This formulation allows us to construct constellations that enforce different aspects of the cooperative transport task. In the following, we define two specific choices: a constellation anchored to the robot base to encourage global alignment with the payload, and a constellation anchored to the end-effector to regulate local contact behavior.

\textbf{End-Effector Contact Constellation.} For each arm contact pad and its designated contact frame ${}^bT_{\text{cf}}^{(r)}$, we define corresponding constellations. As illustrated in Fig~\ref{fig:constellation-reward}, for robot $r$ we spawn five \textit{colinear} points $P^{(r)}_{\text{pad}}$ spanning $50cm$ uniformly along the arm's contact pad normal and corresponding colinear points $P^{(r)}_{\text{cf}}$ spanning the contact frame normal. Matching these constellations enforces both position alignment and surface normal consistency, but allows flexibility in tangential orientations. The associated reward term is
\[
R_{\text{contact}} = - d_{\text{con}}\left(P^{(r)}_{\text{pad}},P^{(r)}_{\text{cf}}\right) 
\]
which drives the contact pad and contact frame constellations to match, effectively emulating rigid attachment.

\textbf{Base Tracking Constellation.} To enforce the robot body to maintain consistent alignment with the payload during transport we define a constellation of three \textit{non-collinear} points on the robot base $P^{(r)}_{\text{base}}$, chosen to resemble adjacent corners of a cube of size $(10 \times 10 \times 10)\,\text{cm}$ (see Fig~\ref{fig:constellation-reward}). The target constellation of three corresponding points $P^{(r)}_{\text{rigid}}$ are defined to represent the target positions of $P^{(r)}_{\text{base}}$ under the assumption that the robot is attached through a rigid kinematic chain through the arm. In particular, the points in $P^{(r)}_{\text{rigid}}$ are computed at each timestep to be the positions that the points in $P^{(r)}_{\text{base}}$ should be given the current payload target command. The base tracking reward is then defined as:
\[
R^{(r)}_{\text{track}} = - d_{\text{con}}\left( P^{(r)}_{\text{base}}, P^{(r)}_{\text{rigid}}\right)
\]
This encourages base motion that is kinematically consistent with the commanded payload motion.

Together, these terms define the overall constellation reward for robot $r$ (Fig.~\ref{fig:constellation-reward}), 
\[
R^{(r)}_{\text{constellation}} = R^{(r)}_{\text{contact}} + R^{(r)}_{\text{track}}.
\]
\subsubsection{Task-Specific Tracking Rewards and Regularization}
Beyond the constellation formulation, which unifies position and orientation alignment into a single smooth objective, we include targeted rewards for specific aspects of the transport task. During contact establishment, a binary \textbf{contact reward} encourages initial surface engagement. For payload control, we track \textbf{height accuracy} during lift/move phases 
% ($R_{\text{height}} = -\|h_{\text{actual}} - h_{\text{commanded}}\|^2$) 
and \textbf{velocity tracking} during transport. These rewards are implemented via exponential kernels over the tracking errors. 
% ($R_{\text{velocity}} = -\|\mathbf{v}_{\text{actual}} - \mathbf{v}_{\text{commanded}}\|^2$).

% Smaller version
To stabilize training and encourage smooth policies, we include regularization terms. As shown in Figure~\ref{fig:reward_timeline}, these penalties are selectively activated based on training phase and episode timing. These include \textbf{torque and joint motion penalties} to prevent excessive actuator effort and high velocities, \textbf{action smoothness penalties} to discourage abrupt control changes, \textbf{phase-specific constraints} such as leg motion penalties during contact formation (seconds 0-4). Additional safety terms include \textbf{levelness penalties} to prevent payload tilting, \textbf{box acceleration penalties} during transport phases and outside-range penalties to constrain robots within assigned contact regions.  These auxiliary terms provide essential regularization while the constellation and tracking rewards drive the primary learning objectives of rigid contact and coordinated transport. Full details of all reward terms are in the extended paper.

% Lengthier version
% Moreover to stabilize training and encourage smooth policies, we include regularization terms that work alongside the constellation and tracking rewards. As shown in Figure~\ref{fig:reward_timeline}, these penalties are selectively activated based on training phase and episode timing:

% \begin{itemize}
%     \item \textbf{Torque penalties} prevent excessive actuator effort throughout all phases
%     \item \textbf{Joint motion penalties} discourage excessively high joint velocities or accelerations, improving stability and reducing mechanical stress
%     \item \textbf{Action smoothness penalties} discourage abrupt control changes that could destabilize contact
%     \item \textbf{Leg motion penalties} are active during pinch and lift phases (seconds 0-4) to focus learning on arm control when base motion is unnecessary
%     \item \textbf{Box acceleration penalties} are applied during lift and move phases to ensure smooth, coordinated motion without oscillations
%     \item \textbf{Lost contact penalty} maintains stable manipulation once contact is established
%     \item \textbf{Levelness penalty} prevents excessive payload tilting by penalizing deviations from horizontal orientation
%     \item \textbf{Outside-range penalty} constrains robots to stay within assigned contact regions
% \end{itemize}

% These auxiliary terms provide essential regularization while the constellation and tracking rewards drive the primary learning objectives of rigid contact and coordinated transport. Full details of all reward terms are in the extended paper. 

\begin{figure*}[t]
    \centering
    \includegraphics[width=0.8\textwidth]{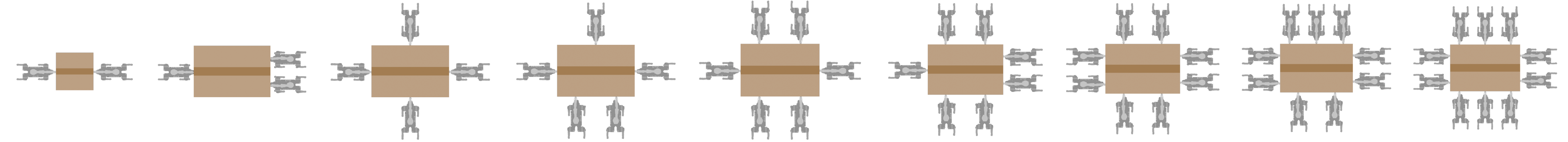}
    \caption{\small Robot arrangements with different team sizes. Robots are evenly distributed around the box 
    ($1.0 \times 1.5 \times 0.7$\,m, mass $2$\,kg), as used in the experiments of 
    Fig.~\ref{fig:comparing_box_info_and_constellation_methods}. For the two-robot case, a smaller box 
    ($0.5 \times 0.4 \times 0.7$\,m) is used. 
    Robot indexing, when referenced, starts from the \emph{top-right corner} and increases clockwise around the box.}
    \vspace{-1em}
    \label{fig:n-robot-arrangments-in-box}
\end{figure*}

\subsection{Training Curriculum}
\label{sec:training_curriculum}

The complexity of coordinated manipulation necessitates a carefully designed curriculum that progressively increases the difficulty of the learning problem. We structure this by carefully defining the generation of training episodes and incrementally adding command modalities to the training phases, as shown in Figure~\ref{fig:reward_timeline}. 

\textbf{Training episode generation.} Training uses a box payload with contact frames $\{{}^{b}T_{\text{cf}}^{(r)}\}$ randomly sampled on its surface subject to force-closure feasibility. Robots start near their assigned frames 
% with small random offsets 
so that arms can reach without base motion, enabling early episodes to focus on contact formation.  

To support robustness and sim-to-real transfer, we apply domain randomization over robot dynamics, contact parameters, payload mass (0.1-2 kg), external forces, initial poses, and sensor/action noise (details in the extended paper).  

We train both controllers with persistent contact-frame observations and controllers where pose information is only given for the first 2 s. Directly training the latter proved difficult, so we begin with full 50 Hz updates and progressively anneal the update rate (25 Hz → 5 Hz → 0.25 Hz → 0 Hz), which yields faster and more stable learning.

\textbf{Training Phase 1: Contact Formation (Pinch).}
In this phase, we limit the command pool for the velocity and the height of the box to be 0, i.e., $\mathcal{C}_{\text{obj}}=(0,0,0,0)$, hence the robots learn only to establish and maintain rigid contact at assigned contact frames. Here, the reward is designed in such a way that the robot is incentivized to reach the box contact surface within 4 seconds - hence, we limit the episode length to 7 seconds for this training phase. Moreover, for this phase, the box mass is set to a very large value (100 kg) to prevent premature motion, and leg movement is heavily penalized since locomotion is unnecessary. This phase focuses purely on arm control and contact regulation.

\textbf{Training Phase 2: Coordinated Lifting (Pinch and Lift).}

In this phase we extend the command pool to include payload height commands in addition to the pinch setting. Thus $\mathcal{C}_{\text{obj}}=(0,0,0,h_{\text{obj}})$ where $h_{\text{obj}} \geq 0$, so the policy must handle both cases: continuing contact maintenance with $h_{\text{obj}}=0$ (as in Phase 1) and executing vertical lifting when $h_{\text{obj}}>0$. Planar velocities remain zero. The box mass is randomized from 0.1 to 2 kg and the episode length is extended to 14 seconds. Robots receive vertical lifting commands and must coordinate to lift the payload while maintaining stable contact, with emphasis on smooth, synchronized motion without oscillation.

\textbf{Training Phase 3: Full Transport (Pinch, Lift and Move).}
In this phase, we employ the complete command pool, which includes nonzero planar velocity as well as height commands, i.e., $\mathcal{C}_{\text{obj}}=(v_{\text{obj}}^x, v_{\text{obj}}^y, \omega_{\text{obj}}^z, h_{\text{obj}})$ with each component permitted to be either zero (as in earlier phases) or nonzero. This command modality requires full coordination of base locomotion and arm control. The episode length remains 14 seconds, but leg movement penalties are removed since robots must demonstrate pinch-lift-move capabilities.

% ╔══════════════════════════════════════════════════════════════════════════════╗
% ║                           SECTION: Experiments and Results                   ║
% ╚══════════════════════════════════════════════════════════════════════════════╝

% \begin{figure*}[t]
%     \centering
%     \includegraphics[width=\textwidth]{figures/comparing_box_masses.pdf}
%     \caption{\small Performance of \ours{} and \oursp{} on different team sizes under varying object masses from 2 to 20 kg, with curve shading indicating mass progression from lighter (2 kg) to heavier (20 kg) objects. For details see Subsection~\ref{subsec:object-mass-scaling}.}
%     \vspace{-1em}
%     \label{fig:comparing_box_masses}
% \end{figure*}

%\vspace{-0.5em}
\section{Experiments and Results}
\label{sec:experiment}

We evaluate our approach through experiments in sim and demonstrations of sim-to-real transfer. Specifically, we examine: (1) the importance of the constellation reward, (2) robustness to limited contact frame information, (3) generalization across different team sizes and payload masses, and (4) transfer to out-of-distribution payloads.

%We evaluate our approach across 3 key dimensions: (1) the importance of the constellation reward, (2) robustness to limited contact frame information, (3) generalization to different team sizes and object masses, and (4) transfer to real-world object geometries.

\subsection{Experimental Setup}

We conduct experiments in simulation with teams of $N \in \{2, \ldots, 10\}$ quadruped-arm robots cooperatively transporting payloads. 
Our primary experiments use a box as the payload with dimensions $1.0 \times 1.5 \times 0.7$\,m and mass 2\,kg with the exception of using a slightly smaller box ($0.5 \times 0.4 \times 0.7$\,m) for $N=2$. Figure~\ref{fig:n-robot-arrangments-in-box} shows how different numbers of robots are positioned around the box. 
%robot placement around the box, with indexing beginning at the top-right corner and increasing clockwise. 

Each evaluation episode is run for 14 seconds with each experimental configuration being run over 500 episodes. Episodes are initialized with robots near their assigned contact frames with small random offsets to test robustness under positioning uncertainty. Episodes then proceed with randomized commands: box velocities $v_x, v_y \sim \mathcal{U}[-0.4, 0.4]$ m/s, angular velocity $\omega_z \sim \mathcal{U}[-0.4, 0.4]$ rad/s, and height $h \sim \mathcal{U}[0.1, 0.3]$ m above ground. 

%We use a box with dimensions $1.0 \times 1.5 \times 0.7$\,m and mass 2\,kg; \emph{for the two-robot case only}, we use a slightly smaller box ($0.5 \times 0.4 \times 0.7$\,m) to avoid excessive size for two robots. For real-world object scenarios, we also evaluate a wooden log (2\,kg, 2 robots), a barrel (3\,kg, 3 robots), and a couch (10\,kg, 5 robots). For each episode, robots are initialized near their assigned contact frames with small random offsets to test robustness under realistic positioning uncertainty. 

To ensure robustness under realistic deployment conditions, all evaluations are performed with dynamics randomization and observation noise enabled. Gaussian noise is injected into all observation channels used by the high-level policy $\pi^h_\theta$ and low-level policy $\pi^b_\theta$, including proprioceptive state $s^{(r)}$, contact-frame pose ${}^{b}T_{\text{cf}}^{(r)}$, and the temporal synchronization signal $t^{\text{sync}}$. These perturbations simulate sensing imperfections and environmental variability.
%, ensuring that reported performance reflects robustness comparable to real-world scenarios.

We evaluate performance using five metrics: (i) linear velocity tracking error (RMS error in $x$ and $y$ directions), (ii) angular velocity tracking error about $z$-axis, (iii) height tracking error, (iv) drop rate (percentage of episodes where the payload is dropped or fails to lift), and (v) robot failure rate (percentage of episodes with robot collisions or falls).

\subsection{Model Variants}
To evaluate our key design choices, we focus on two questions: (1) does the constellation reward improve collaborative PLM performance, and (2) how important is continual access to contact frame information versus only at initialization? To answer these, we train four model variants that cross reward formulation ($\text{const}^+$ vs.\ $\text{const}^-$) with contact frame observability ($\text{cf}^+$ vs.\ $\text{cf}^\text{init}$), denoted as
\[
\text{decPLM}(X, Y), \;\; X \in \{\text{const}^+, \text{const}^-\}, \; Y \in \{\text{cf}^+, \text{cf}^{\text{init}}\}.
\]
By default all decPLM controllers are trained with 2 robots. 

Since no prior methods address the decentralized PLM problem, we compare against three baselines. First, a 1-robot system with no payload provides an idealized reference for linear velocity tracking, angular velocity tracking, and fall rate. Second, we train with three robots, which we denote by $\text{decPLM}_{\text{3r}}$. In later experiments, we evaluate a centralized architecture to assess the impact of decentralization.

%Table~\ref{tab:ours-variants} summarizes these variants. We introduce shorthand notation for all methods: \ours{} (Init-Pose + Constellation), \oursp{} (Full-Pose + Constellation), and baseline non-constellation variants \base{} and \basep{}.

%This design allows us to answer key questions: How important is the unified reward formulation compared to traditional separate position/orientation losses? Can the approach maintain performance when box pose information is masked after initial contact formation?

\begin{figure*}[t]
    \centering
    \includegraphics[width=\textwidth, trim=0cm 0.2cm 0cm 0.4cm,clip]{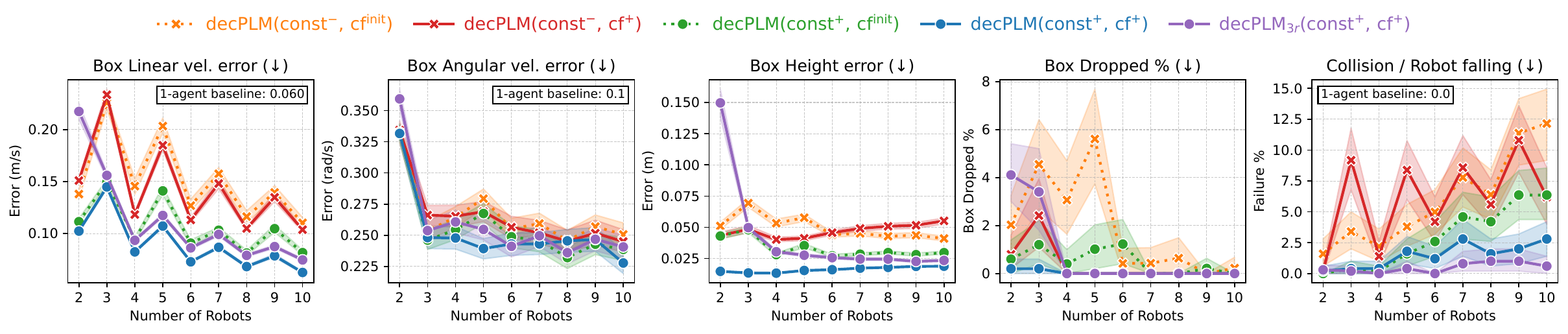}
    \caption{\small We compare performance across different team sizes for four model variants:
    \base{}, \basep{}, \ours{}, and \oursp{}.
    Line style indicates access to contact-frame pose (solid = cf$^+$ vs. 
    dotted = cf$^{\text{init}}$), marker shape indicates use of the constellation reward
    (circle = const$^+$ vs. cross = const$^-$, and colors correspond
    to the methods as listed. We also include a \textbf{1-robot baseline}, which does not
    carry any load but is provided as an idealized reference value. Finally, $\text{decPLM}_{\text{3r}}$ is a variant trained with three robots instead of two.  For details, see Subsection~\ref{subsec:consellation-reward-effectiveness} and~\ref{subsec:object-mass-scaling}.
    }
    \vspace{-1.1em}\label{fig:comparing_box_info_and_constellation_methods}
\end{figure*}

\begin{figure}[t]
    \centering
    \includegraphics[width=0.8\columnwidth]{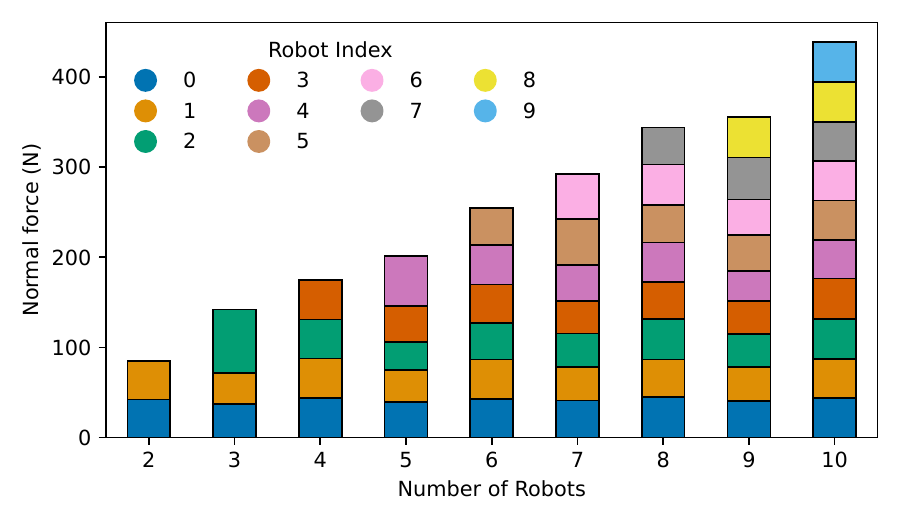}
    \caption{\small The figure shows the average normal force exerted on the box by each robot after the lifting phase for \ours. For details see second paragraph on Subsection~\ref{subsec:team-size-generalization-force-distribution}.}
    \vspace{-1.1em}
\label{fig:contact_force_exerted_by_each_agent_after_lifting_phase}
\end{figure}

% \begin{table}[h]
% \centering
% \small
% \resizebox{\columnwidth}{!}{%
% \begin{tabular}{l c c}
% \toprule
% \textbf{Variant} & \makecell{\textbf{Constellation}\\\textbf{reward}} & \makecell{Contact Frame\\Pose (\textbf{${}^bT_{\text{cf}}$})\\\textbf{available after $2s$}} \\
% \midrule
% \ours{} (Init-Pose + Constellation) & \cmark & \xmark \\
% \oursp{} (Full-Pose + Constellation) & \cmark & \cmark \\
% \base & \xmark & \xmark \\
% \basep & \xmark & \cmark \\
% \bottomrule
% \end{tabular}%
% }
% \caption{\small Summary of method variants compared in our experiments. 
% A \cmark\ under \emph{Constellation reward} indicates use of our unified reward; 
% \xmark\ means separate position and orientation reward terms. 
% A \cmark\ under ${}^bT_{\text{cf}}$ available means the contact-frame pose is continuously available throughout execution, 
% while \xmark\ means it is provided only during the first 2 seconds for contact formation.}
% \vspace{-1em}
% \label{tab:ours-variants}
% \end{table}

\vspace{-0.25em}
\subsection{Constellation Reward Effectiveness}
\label{subsec:consellation-reward-effectiveness}
Figure~\ref{fig:comparing_box_info_and_constellation_methods} compares performance across the model variants for teams of 2-10 robots. The constellation-based methods substantially outperform traditional approaches. Both \ours{} and \oursp{} achieve lower tracking errors and drop rates compared to their non-constellation counterparts across all team sizes.

Surprisingly, by comparing \ours{} to \basep{} we see that the constellation reward is more critical to performance than continuous contact frame information. 
For example, with 5 robots, \ours{} is approximately 50\% better at tracking velocity despite the contact pose information being masked. This indicates that a constellation reward structure enables robust contact maintenance even when explicit contact frame information is unavailable after the pinch phase.

\vspace{-0.25em}
\subsection{Team Size Generalization and Force Distribution}
\label{subsec:team-size-generalization-force-distribution}

Our policies, trained exclusively with 2-robot, generalize effectively to larger configurations without retraining. Figure~\ref{fig:comparing_box_info_and_constellation_methods} shows that performance improves consistently as team size increases: linear velocity errors decrease from 0.1 to 0.02 m/s when scaling from 2 to 10 robots, which is an 80\% drop in error, while box drop rates fall from ~5\% to less than 1\%. This improvement stems from better force distribution and increased redundancy provided by additional robots, as evidenced by our force below. 

Comparing to $\text{decPLM}_{\text{3r}}(\text{const}^+,\text{cf}^+)$ (i.e. 3-robot training), we find 2-robot training performs as well or better across team sizes, with only a slight improvement in collision/fall rate for 3-robot training. One anomaly occurs where the 3-robot model performs worse on 3-robot teams, warranting further study, but overall there is little evidence that 3-robot training justifies its added cost. We also observe a small increase in linear velocity error (and, to a lesser extent, other metrics) when moving from even to odd team sizes. This is due to geometric asymmetry in contact placement: even teams allow approximately opposing force pairs, whereas odd teams require uneven moment balancing. The learned policy compensates by redistributing normal forces (Figure \ref{fig:contact_force_exerted_by_each_agent_after_lifting_phase}), but this introduces slightly higher tracking error.

We also analyze how contact forces are distributed across team members using \ours{}. As shown in Figure~\ref{fig:contact_force_exerted_by_each_agent_after_lifting_phase}, in asymmetric team configurations - e.g., 3-robots with one robot on one side and 2 on the other (see Fig~\ref{fig:n-robot-arrangments-in-box} for team arrangement) - the co-located robots together apply approximately the same total force as the single robot on the opposite side, indicating that the policy generalizes well by balancing forces despite geometric asymmetry. In general, we observe this asymmetric balancing behavior most clearly in smaller teams, but as the number of robots increases, individual force contributions become more uniform, indicating that larger teams distribute loads more evenly and rely less on compensating for geometric asymmetry.

\subsection{Payload Mass Scaling}  
\label{subsec:object-mass-scaling}

\begin{figure*}[t]
    \centering
    \includegraphics[width=\textwidth, trim=0cm 0.2cm 0cm 0.4cm,clip]{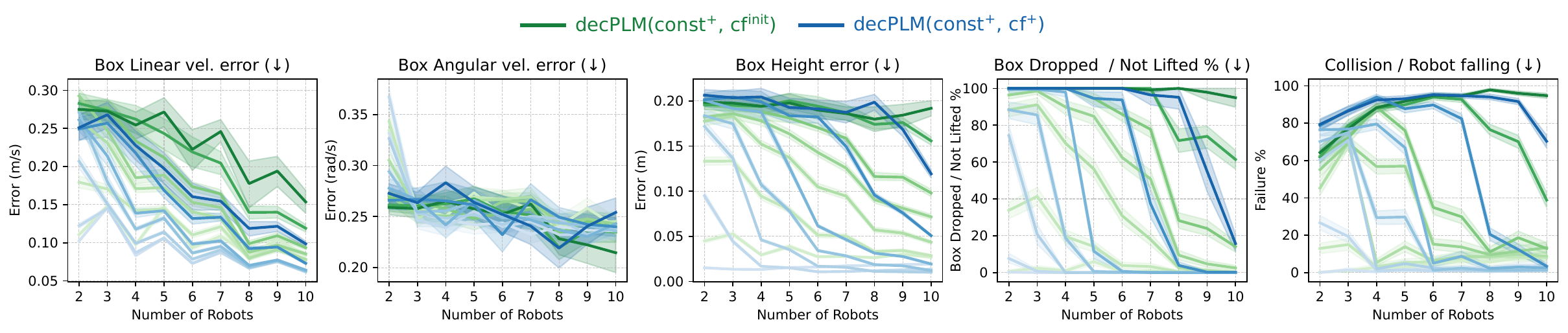}
    \caption{\small Performance of \ours{} and \oursp{} on different team sizes under varying payload masses from 2 to 20 kg, with curve shading indicating mass progression from lighter (2 kg) to heavier (20 kg) payloads. For details see Subsection~\ref{subsec:object-mass-scaling}.}
    \vspace{-1em}
    \label{fig:comparing_box_masses}
\end{figure*}

Figure~\ref{fig:comparing_box_masses} evaluates performance under varying payload masses from 2 to 20 kg, with curve shading indicating mass progression. For the \ours{} variant - without continuous pose information, additional robots generally improve performance by providing better force distribution for heavier payloads. However, a critical threshold emerges around 15 kg: beyond this point, even larger teams struggle with the substantial normal forces required for lifting, leading to elevated drop rates and robot failures.

Including pose information significantly improves performance: \oursp{} achieves consistently lower velocity errors across all masses compared to \ours{}. This gap becomes dramatic at 15 kg, where \oursp{} maintains an 18\% drop rate and 5\% failure rate, while \ours{} degrades sharply to 60\% drop rate and 40\% failure rate. These results indicate that for heavy payloads requiring precise force coordination, continuous knowledge of contact frame pose becomes crucial for our current approach to maintaining system stability and performance.

\begin{table}[t]
\centering
\small
\setlength{\tabcolsep}{3pt}
\renewcommand{\arraystretch}{0.95}
\resizebox{\columnwidth}{!}{%
\begin{tabular}{C{16mm} C{8mm} l c c c c c}
\toprule
\makecell{\textbf{Scene}} & \makecell{\textbf{\rotatebox{90}{Robots}}} & \makecell{\textbf{Methods}} & \makecell{Linear\\vel.\\error} & \makecell{Ang\\vel.\\error} & \makecell{Height\\error} & \makecell{Payload\\Dropped\\\%} & \makecell{Robot\\falling\\\%} \\
\midrule
\multirow{5}{*}{\raisebox{-0.5\height}{\includegraphics[height=8mm]{\detokenize{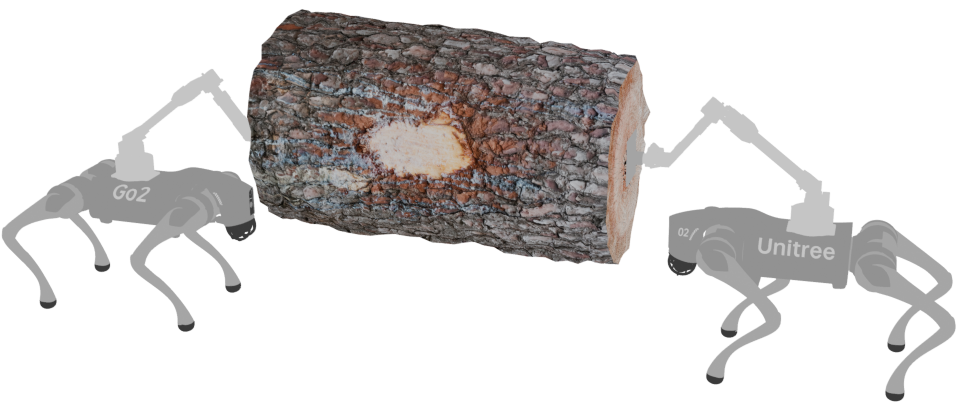}}}} & \multirow{5}{*}{2} & \oursp & 0.103 & 0.290 & 0.017 & 0.3 & 0.0 \\
 &  & \ours & 0.117 & 0.278 & 0.049 & 2.6 & 0.0 \\
 &  & \textcolor{teal}{$\text{decPLM}_{\text{3r}}(\text{const}^+,\text{cf}^+)$} & 0.134 & 0.279 & 0.071 & 12.2 & 0.0 \\
 % &  & \textcolor{teal}{decentralized-3-robots (cf init)} & 0.130 & 0.289 & 0.069 & 9.600 & 0.000 \\
 &  & \textcolor{violet}{specialized-wooden-log} & 0.099 & 0.288 & 0.052 & 0.0 & 0.0 \\
 % &  & \textcolor{violet}{specialized-wooden-log (cf init)} & 0.105 & 0.296 & 0.022 & 2.300 & 0.000 \\
 &  & \textcolor{darkgray}{centralized-2-robots} & 0.119 & 0.278 & 0.056 & 6.7 & 0.4 \\
\midrule
\multirow{4}{*}{\raisebox{-0.5\height}{\includegraphics[height=12mm]{\detokenize{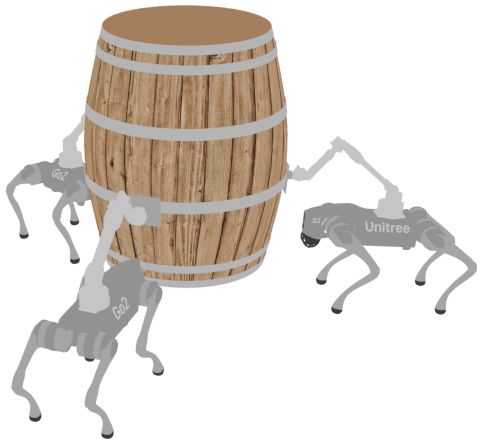}}}} & \multirow{4}{*}{3} & \oursp & 0.184 & 0.293 & 0.031 & 0.1 & 0.0 \\
 &  & \ours & 0.135 & 0.310 & 0.080 & 6.0 & 0.2 \\
 &  & \textcolor{teal}{$\text{decPLM}_{\text{3r}}(\text{const}^+,\text{cf}^+)$} & 0.116 & 0.278	& 0.046 &	1.0 & 0.0 \\
% &  & \textcolor{teal}{decentralized-3-robots} & 0.126 & 0.292 & 0.087 & 6.600 & 0.200 \\
 &  & \textcolor{violet}{specialized-barrel} & 0.097 & 0.271 & 0.021 & 0.1 & 0.0 \\
 % &  & \textcolor{violet}{specialized-barrel (cf init)} & 0.112 & 0.298 & 0.073 & 1.300 & 0.200 \\
\midrule
\multirow{4}{*}{\raisebox{-0.5\height}{\includegraphics[height=10mm]{\detokenize{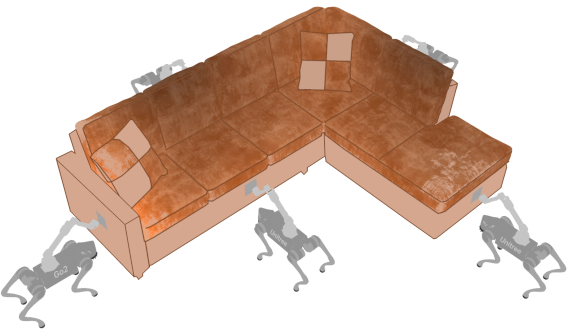}}}} & \multirow{4}{*}{5} & \oursp & 0.145 & 0.229 & 0.098 & 15.0 & 0.4 \\
 &  & \ours & 0.264 & 0.268 & 0.187 & 51.5 & 8.8 \\
 &  & \textcolor{teal}{$\text{decPLM}_{\text{3r}}(\text{const}^+,\text{cf}^+)$} & 0.187 & 0.225	& 0.171 & 43.6 & 1.8 \\
% &  & \textcolor{teal}{decentralized-3-robots} & 0.210 & 0.241 & 0.185 & 49.600 & 3.400 \\
  &  & \textcolor{violet}{specialized-couch} & 0.178 & 0.256 & 0.091 & 23.0 & 2.4 \\
 % &  & \textcolor{violet}{specialized-couch (cf init)} & 0.182 & 0.257 & 0.094 & 24.100 & 2.200 \\
\midrule
\bottomrule
\end{tabular}%
}
\caption{\small Comparing performance on out-of-distribution payloads. See Subsection~\ref{subsec:realworld-scene} for details.}
\vspace{-1.5em}
\label{tab:comparing_different_objects}
\end{table}

\subsection{Out-of-Distribution Payloads and Alternative Baselines}
\label{subsec:realworld-scene}

We next evaluate generalization to out-of-distribution payloads and compare against alternative baselines. Test payloads include a wooden log (2 kg, 2 robots), a barrel (3 kg, 3 robots), and a couch (10 kg, 5 robots) (Table~\ref{tab:comparing_different_objects}). 

We compare against two additional baselines trained with persistent observability ($\text{cf}^+$) and constellation reward ($\text{const}^+$): (1) \emph{centralized control}, with joint actions from global state (practical only up to 2 robots), and (2) task-specialized decentralized controllers trained directly on each payload rather than the default box. Results in Table~\ref{tab:comparing_different_objects}.

% We now consider how well the trained PLM controllers generalize to out-of-distribution payloads and how they compare to alternative architectures and specialized training on these payloads. We evaluate on three new payloads: a wooden log (2\,kg, 2 robots), a barrel (3\,kg, 3 robots), and a couch (10\,kg, 5 robots) as shown in Table ~\ref{tab:comparing_different_objects}.

% We compare to several alternative approaches: (1) \emph{centralized control}, where the policy has access to the global state of all robots and jointly outputs actions for all robots. We found that it was only practical to train this system to reasonable performance for 2 robots. (2) team-size-specific training where policies were trained for the exact evaluation team size rather than our default training with just 2 robots, and (3) task-specialized decentralized controllers trained specifically for each evaluation payload rather than our default box payload. Results are shown in Table~\ref{tab:comparing_different_objects}.

%\noindent\textit{The results reveal several key insights:}
\textbf{Geometric complexity increases information requirements:} As payloads become more irregular, continuous contact-frame pose information (\oursp{}) becomes increasingly important, with the drop-rate gap relative to \ours{} growing from 2.3\% (log) to 5.9\% (barrel) and 33.5\% (couch).

% \textbf{Geometric complexity increases information requirements:} As payloads become more irregular (log $\rightarrow$ barrel $\rightarrow$ couch), continuous contact frame pose information (\oursp{}) becomes increasingly important for maintaining low drop rates and stable tracking. This is evident by the drop rate gap of 2.3\%, 5.9\%, and 33.5\% between \oursp{} and \ours{} for log, barrel, and couch, respectively. 

\textbf{Constellation reward enables strong generalization:} Our general approach matches or exceeds payload-specific controllers, demonstrating effective transfer from box training to diverse geometries at much lower computational cost.

% \textbf{Constellation reward enables competitive generalization:} Our general approach matches or outperforms payload-specific controllers, showing effective transfer from box training to diverse geometries. While specialized policies might catch up with longer training, this requires substantially higher computational cost for the 3 and 5 robot teams.

\textbf{Centralized control offers limited benefit:} In the two-robot setting, centralized execution does not outperform decentralized control, likely because the added dimensionality complicates optimization while CTDE already supplies strong coordination signals.

% \textbf{Centralized control shows no clear advantage:} Although the centralized controller has access to global state, it does not outperform the decentralized variant in the two-robot setting. The increased input-output dimensionality appears to make optimization more difficult, and the centralized critic in CTDE already provides strong coordination signals during training. For modest team sizes, explicit centralization at execution offers limited additional benefit.

%centralized methods don't consistently outperform our decentralized approach, particularly on simpler geometries where the constellation reward provides sufficient coordination.

\textbf{Training team size does not determine generalization:} Policies trained with 2 robots generally match or outperform those trained with 3, suggesting that domain randomization with 2 robots already provides sufficient diversity for broad generalization.

\vspace{-0.5em}
\section{Real-World Demonstration}
\label{sec:sim-to-real}
We evaluated \ours{} on teams of two, three and four Unitree Go2 robots with D1 arms, replacing the stock end-effectors with custom rubber-covered contact pads. For safety, experiments used lightweight boxes (1-2kg).

Real-world deployment revealed key challenges: unknown arm dynamics requiring manual gain tuning, joint encoder offsets needing external calibration, limited arm torque restricting payload capacity, and box deformability disrupting contact cues. Despite these issues, the policy executed coordinated pinch--lift--move sequences (Fig.~\ref{fig:lead_image}), demonstrating that constellation-based coordination transfers to hardware, though with a substantial sim-to-real gap due to current hardware and calibration limitations.

\section{Summary and Future Work}
\label{sec:conclusion}
% \vspace{-0.5em}

We introduced the cooperative pinch--lift--move (PLM) problem for ungraspable objects and presented \textbf{decPLM}, a decentralized learning framework that requires neither communication nor rigid coupling. By combining a locomotion-manipulation hierarchy with a constellation reward that promotes rigid-like contact behavior, the method enables a shared policy to generalize across team sizes and payloads. Simulations show that this reward is critical for coordination and that policies trained with two robots transfer to larger teams and varied payloads. Real-world experiments with two to four robot teams validate the approach while exposing practical limitations, including torque limits, calibration, and object deformability. Future work will explore high-level planning, autonomous contact assignment, richer sensing, more varied terrain, and extension to other robot platforms.

\bibliographystyle{ieeetr}
% \vspace{-1em}
\bibliography{references}

\end{document}